%% file: acl_latex.tex
\pdfoutput=1

\documentclass[11pt]{article}
\usepackage[final]{acl}

\usepackage{times}
\usepackage{latexsym}
\usepackage[T1]{fontenc}
\usepackage[utf8]{inputenc}

\usepackage{amsfonts}
\usepackage{amsmath}
\usepackage{amssymb}        
\usepackage{bold-extra}     
\usepackage{bm}             

\usepackage{graphicx}       
\usepackage{multirow}       
\usepackage{booktabs}       
\usepackage{enumitem}       
\usepackage{subcaption}     
\usepackage[hang,flushmargin]{footmisc}  

\usepackage{float}

\title{Improving Narrative Relationship Embeddings by Training with Additional Inverse-Relationship Constraints}

\author{Mikolaj Figurski \\
  Department of Computer Science \\
  Department of Economics \\
  Emory University \\
  Atlanta, GA, 30322, USA \\
  \texttt{mfigurs@emory.edu} \\}

\begin{document}
\maketitle

\input{tex/abstract}
\input{tex/introduction}
\input{tex/related-work}
\input{tex/approach}
\input{tex/experiments}
\input{tex/analysis}
\input{tex/conclusion}
\input{tex/acknowledgements}

\bibliography{custom}

\cleardoublepage\appendix
\input{tex/appendix}

\end{document}

%% file: tex/abstract.tex
\begin{abstract}
We consider the problem of embedding character-entity relationships from the reduced semantic space of narratives, proposing and evaluating the assumption that these relationships hold under a reflection operation. We analyze this assumption and compare the approach to a baseline state-of-the-art model with a unique evaluation that simulates efficacy on a downstream clustering task with human-created labels. Although our model creates clusters that achieve Silhouette scores of -.084, outperforming the baseline -.227, our analysis reveals that the models approach the task much differently and perform well on very different examples. We conclude that our assumption might be useful for specific types of data and should be evaluated on a wider range of tasks.
\end{abstract}

%% file: tex/introduction.tex
\section{Introduction}
\label{sec:introduction}

Of the many different types of ideas humans care for enough to pick up and read text, perhaps the most fun has got to be stories. There are many practical examples -- things like history, backstories for concepts, and illustrative narratives help us understand the world around us and the cultures of the people we share it with. Teaching computers to understand this format is no mean feat. In this paper, we focus specifically on creating a model that takes examples of a narrative and dialogue format and embeds them into a lower-dimensional vector space that accurately captures semantic similarities. If successful, this work can help by improving methods of constructing knowledge graphs, especially such that focus on character relationships, thereby aiding further classification or knowledge mining problems. Furthermore, improving these techniques could allow us to easily discover character archetypes by performing clustering tasks on final embeddings.

State-of-the-art (henceforth, SOTA) models have proved successful in various types of relation extraction, but have rarely been applied to narratives and character-relationship extraction, a task which may pose a unique challenge. We propose that an additional assumption about the inverses of these relationships could increase the space of our learning to make it more effective. To the best knowledge of the author, we will be presenting a new assumption related to inverse relationships. This will then be evaluated by training transformer models using both approaches on a narrative-corpora dataset of Star-Trek scripts, originally compiled by \citet{broughton-2018}, and comparing performance and error types.

We reduce the problem to narrative relationship extraction because it will help us evaluate our method, as discussed further in Section \ref{approach:embed}. The reduced problem involves restricting the Named Entity Recognition (henceforth, NER) process to just character names and thereby presenting a reduced semantic space relative to the one baseline approaches are designed for. The authors hope that this will in turn a) make it easier to analyze where the assumption applies and where it fails, as discussed in Section \ref{sec:analysis}, and b) ensure that the baseline model struggles to some degree and thereby leaves more room for improvement.

%% file: tex/related-work.tex
\section{Related Work}
\label{sec:related-work}

The task of relationship extraction as initially defined by \citet{cardie-1997} has been initially dominated by approaches focused on building relationships from pre-constructed Knowledge Base graphs. A standard baseline test for new model architectures in this space has been proposed by \citet{hao-etal-2018}, and their scoreboard has been dominated, early on, by GNN learning methods.

The process of efficient Knowledge Base (henceforth, KB) Graph construction from a text document has been examined by \citet{clancy-etal-2019}, and, with a focus on unsupervised learning, by \citet{cau-fairbanks-2019}. The latter works to support the SemanticModels Julia tool by parsing conceptual entities from source code and associating them with natural language concepts in an RDF graph, which in turn was built by decomposing text descriptions into <subject, verb, object> triplets, represented by two word embeddings $h, t$ and a transformation vector $r$: $(h, r, t)$. This standard representation has found its way into literature focused on other fields as well.

Early approaches that focus specifically on relationship extraction from graphs include one by \citet{kartsaklis-etal-2018}, proposing multi-sense LSTMs trained on random walks in a KB Graph to predict edges between additional entities. In the same category, \citet{riedel-etal-2013} applies universal schemas and matrix factorization to expand the scope and efficiency of then-current relationship extraction methods. A lot of this work stayed solidly below human-level performance, however, as noted by the aforementioned baseline test created by \citet{hao-etal-2018}. These papers all adopt a standard of representing relationships as vectors mapping between entity embeddings, an idea that has proven critical for the field.

In line with this graph-based approach, several interesting architectures for KB Graphs have been proposed to aid learning in this environment. One potential structure has been outlined by \citet{blin-ines-2022}. Blin models a narrative as a sequence of objective events and describes some of the research questions involved in representing these as knowledge graphs. They propose an ideal representation structure based on the Simple Event Model (SEM) first defined by \citet{hage-etal-2011}, which creates nodes for all events, actors, places, times, and additional constraints, to which Blin added causal and co-participation edges to better reflect narrative (vs just simple event) structures. Relationship extraction by itself, however, could happen by building much simpler structures.

An interesting alternative KB Graph architecture known as Temporal Knowledge Graphs can be built and utilized using CyGNet, a newer method offered by \citet{zhu-etal-2021} which looks at Sequential Copy-Generation Networks to help predict temporal patterns. Their method has two specific inference modes: Copy and Generation, which are combined to build the final result of a query. The unique Copy mechanism is a modified version of a Pointer Network previously used for NLG and allows the network to directly cite content it has seen before. When combined with the aforementioned Generation mode, this mechanism allows them to achieve superior future-fact predictions than state-of-the-art methods that don't take into account temporal facts or recurring patterns. This is a potential avenue for edge prediction with temporal aspects and thereby supporting relationship extraction over long-spanning narratives or other corpora with this structure.

Many of these approaches were supplanted by the advent of BERT, originally presented by \citet{devlin-etal-2019-bert}. Utilizing the context-aware BERT transformer models, \citet{liang-etal-2019} offered KG-BERT for Knowledge Graph completion, which, according to previous methods, could plausibly be adapted to relationship extraction.

However, the current state-of-the-art (SOTA) approach to generic relationship embeddings is offered by Google Research \citet{livio-baldini-etal-2019}, which pre-trains a BERT transformer to learn a constant relationship without the involvement of any intermediary graphs at all, i.e directly from the text. This approach is geared towards detecting generic relationship classes such as \textit{Cause/Effect} or \textit{PartOf} over smaller pieces of text -- with some probabilistic masking that encourages the model to learn relationships based on just the surrounding context. This approach maintains the highest visible score on the test proposed by \citet{hao-etal-2018}.

Finally, for general evaluation methodology, we will draw on the work of \citet{schnabel-etal-2015-evaluation}. In their analysis of approaches to word-embedding evaluation, they identify four broad categories of absolute intrinsic evaluation methods and show how these perform analogously to what humans actually prefer. Their paper was the motivation for our evaluation method, described further in Section \ref{approach:eval}.

%% file: tex/approach.tex
\section{Approach}
\label{sec:approach}

The goal of this study is to propose a new method of training a transformer for narrative relationship extraction, in order to help downstream Knowledge Graph or clustering tasks on narrative corpora. These tasks are a natural fit and have been frequently co-dependent on the problem we are studying, but could be further improved by additional research. The proposed relationship embeddings should allow us to represent character interactions, sentiment, and especially general archetypes in a directed manner (for downstream directed graphs) as described in Section \ref{approach:embed}. In order to simulate effectiveness on these tasks, the relationship embeddings will be tested against state-of-the-art (henceforth, SOTA) approaches by looking at the accuracy and confidence of their grouping in Section \ref{approach:eval}.

Our method is built to make additional assumptions that may improve performance on, specifically, narrative texts containing stories and character interactions. We discuss the dataset we focus on, as well as the unique requirements of this data, the new model, and our approach further in Sections \ref{approach:data} and \ref{approach:embed}.

\subsection{Data}
\label{approach:data}

Our approach utilizes publicly-released scripts for the 4 TV shows and 2 movies of the Star Trek franchise, compiled and formatted by \citet{broughton-2018}. This dataset is not considered at all by relationship-extraction or graph-construction literature but is reflective of the unique downstream tasks our approach is trying to enable. The dataset contains scripts, held as raw text, for each episode in the TV series and in a singular 'episode' for the 2 movies, allowing for an easy and rigorous definition of the data: a list of 6 shows $S = [S^{(1)},\dots, S^{(6)}]$ each comprised of a list of episodes: $S^{(i)} = [C^{(i)}_1, \dots, C^{(i)}_{k^{(i)}}]$, where $k^{(i)}$ is the number of episodes found in the show $S^{(i)}$. For simplicity, we can flatten this structure by ignoring the separation between shows and defining: $ C = \bigoplus_{i=1}^6 C^{(i)} = [C_0, \dots, C_k]$ where $k=\sum_{i=0}^5 k^{(i)}$, i.e the number of episodes across all 6 shows. An individual episode $C_i$ will represent the base level of the corpus, from which we will parse examples for the model.

Although multiple previous approaches need to perform Named Entity Recognition (NER), KB Construction, and/or rely on imperfectly labeled datasets, the format of these corpora allows for the easy extraction of named entities. As seen in Example \ref{ex:example_span}, the corpora follow a clear <ENTITY: Dialogue> pattern. By cleaning each episode $C_i$ of title/afterword text and removing scene transition labels, we can define it to consist of some amount of $D$ Dialogue entries: $C_i = [D^{(i)}_1, \dots, D^{(i)}_{p^{(i)}}]$ for some number of dialogue entries $p^{(i)}$, where each entry has an entity and dialogue $e, d$, such that: $D_i = \langle e_i, d_i \rangle$. As an aside, both of these values can be parsed by splitting on the character ':', and looking for capital letters for the entity -- the regex code we used for this task is included in the final public repository released in Appendix \ref{sec:appendix}.

As described previously, both relationship-extraction \cite{livio-baldini-etal-2019} and Knowledge Graph construction \cite{bordes-etal-2013} literature has defined relationships as translations in embedding space and has focused on the use of triplets with a structure such as <subject, verb, object>. We adopt the notation used by the former, which is more recent and which defines the relationship between two 'label' entities $s_1$ and $s_2$ as consisting of those entities and the surrounding context $x$, as shown in Equation \ref{eq:relationship_vector} from their paper.

\begin{equation}
    \label{eq:relationship_vector}
    r = (x, s_1, s_2)
\end{equation}

In order to match this notation as closely as possible, we can define an individual Example $E$ as 'bi-grams' of the aforementioned individual Dialogue entries, generally: $E_i = \langle e_i + d_i + e_{i+1} + d_{i+1}, e_i, e_{i+1} \rangle = \langle x_i, s_i, o_i \rangle$ wherever both $D_i$ and $D_{i+1}$ exist. Note that in this way, we are defining $e_i$ to be the subject, $e_{i+1}$ to be the object, and their composed dialogue to be the 'relationship' dialogue -- this is the structure we will train our model on, and will represent a basic relation statement. This results in examples such as the one shown in Example \ref{ex:example_span}. We can further define a generic Example list $E = \bigoplus_{i=1}^k C_i = [E_0, \dots, E_p]$ where $p=\sum_{i=0}^k p^{(i)}-1$ is the total number of examples we can build from the dialogue of each corpus.

\begin{figure}[htbp!]
    \textit{
    \underline{KAITAAMA}: you are not to leave this chamber. \\
    \underline{TUCKER}: with all due respect, I'm not one of your subjects.
    }
    \caption{Slightly edgy example of a span from the text, with named entities $s$ and $o$ underlined, as well as some surrounding context $x$.}
    \label{ex:example_span}
\end{figure}

Note that, before generating the examples, the individual dialogues need to be 'deduplicated' to ensure no consecutive dialogues of the same character exist. While evaluating charaters' relationship with themselves is an interesting prospect, the author has concluded it's out of scope for this paper.

\subsection{Narrative Relationship Embeddings}
\label{approach:embed}

This section describes how we aim to build vectors representing Narrative Relationships between named entities from the Examples $E$ gathered in the previous step.

Parsing narrative embeddings poses a different set of challenges compared to generic 'categorical-relationship' extraction as seen by the literature. This paper proposes to build these by training a transformer based on category-relationship parsing offered by \citet{livio-baldini-etal-2019}.

Their approach builds relationship vectors that aim to consistently transform between the representation of unique tokens of matching entity-pairs by learning on text with those entities masked probabilistically, such that the model can focus only on context. Following this technique, we will surround the two core entities in each of our examples with special tokens $\mathrm{[SU]}$, $\mathrm{[/SU]}$, $\mathrm{[OB]}$, and $\mathrm{[/OB]}$, and allow the model to represent relationships by taking the difference of the logits output for the $\mathrm{[SU]}$ and $\mathrm{[OB]}$ tokens.

\begin{figure}[htbp!]
\centering
\includegraphics[width=\columnwidth]{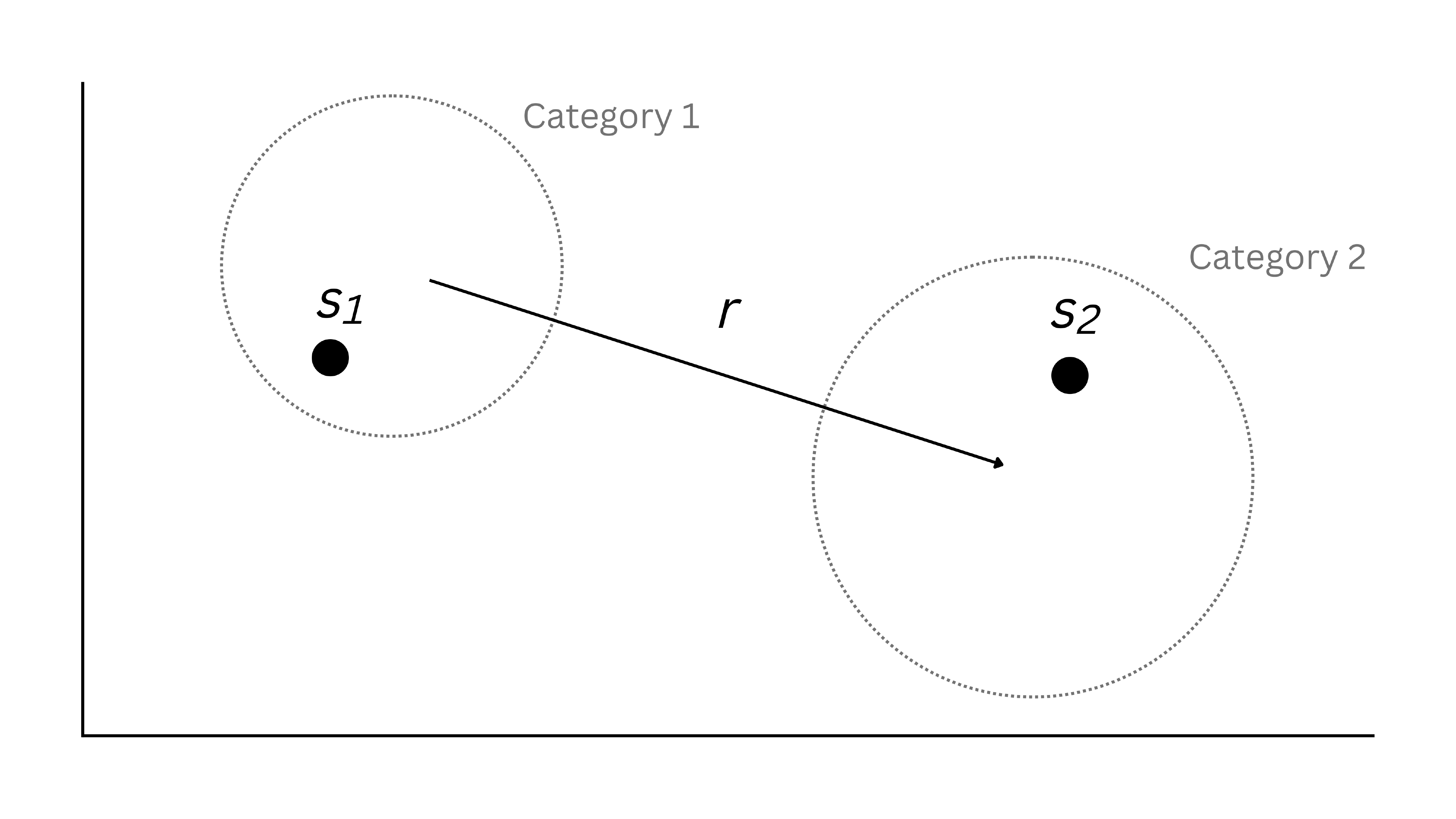}
\caption{Representation of categorical-relationship vector as it relates $s_1$ and $s_2$ across different categories in some semantic space.}
\label{fig:cat_relationship_vector}
\end{figure}

As their paper described, their goal allows this technique to pick up generic, archetypal relations between \textit{categories} of objects. Due to their adoption of the NER technique and admission of all types of possible Named Entities into the training set, their model essentially builds embeddings representing category relationships like ownership, operation, distance, involvement, etc. This results in a transforming vector akin to that represented in Figure \ref{fig:cat_relationship_vector}, which maps from wide categories of Named Entities into other ones based on the surrounding relationship text. Their formal notation for this input, offered in Equation \ref{eq:relationship_vector}, describes the transformation as a function of context $x$ and two label entities $s_1, s_2$.

\begin{figure}[htbp!]
\centering
\includegraphics[width=\columnwidth]{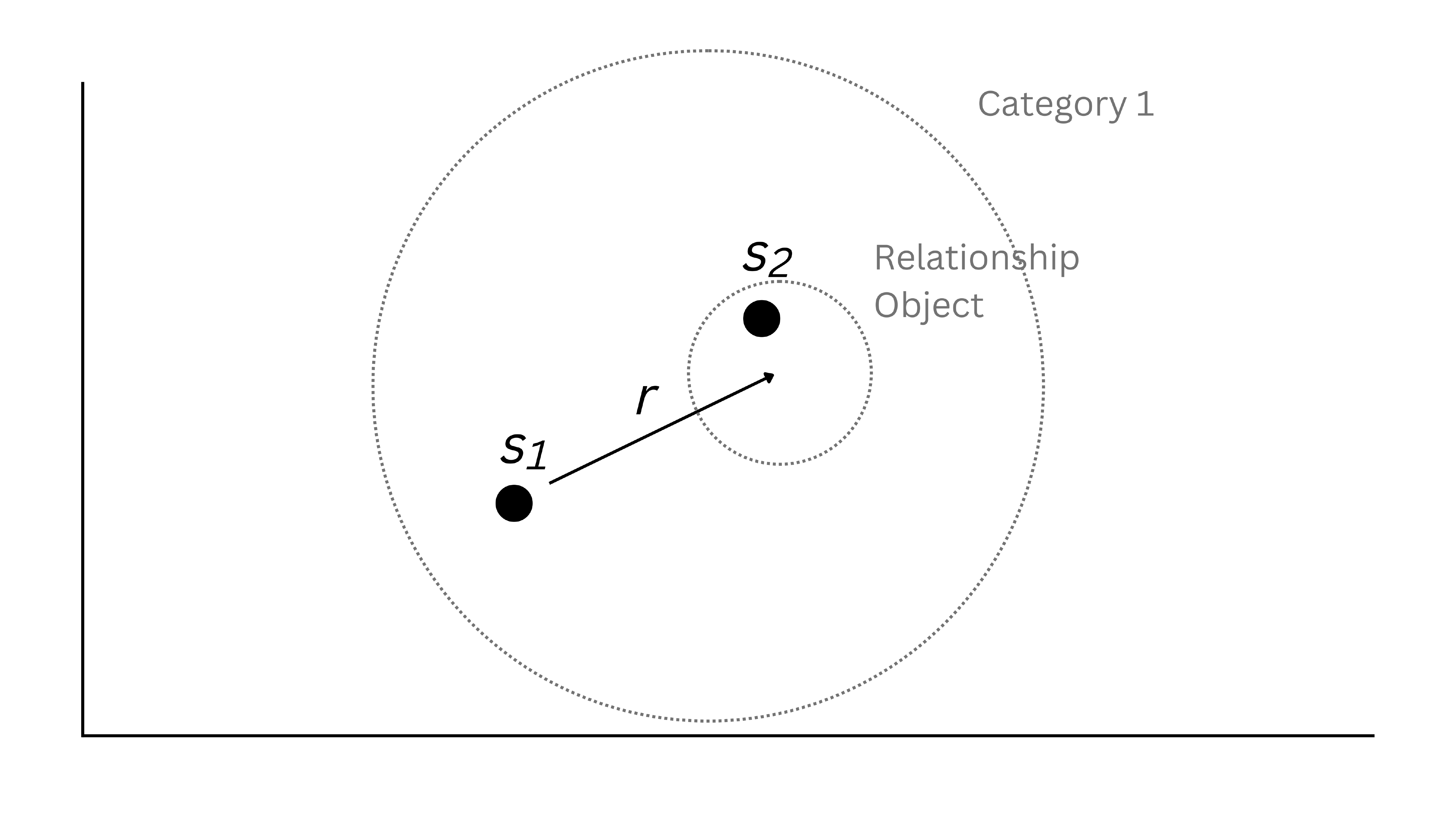}
\caption{Representation of narrative-relationship vector as it relates subject $s_1$ and object $s_2$ within the same 'character' category in a 2D word-embedding space}
\label{fig:nar_relationship_vector}
\end{figure}

Conversely, narrative relationship embeddings are harder to train. Instead of allowing all types of Named Entities that might be picked up by a NER step, the downstream tasks this paper focuses on are only really interested in characters and the relations between those, \textit{making the semantic space much smaller} and therefore harder to train for effectively. This shift is represented by the illustration in Figure \ref{fig:nar_relationship_vector}. It is not obvious how well the baseline approach performs on this reduced space, as all of the listed examples from the presenting paper focus on cross-categorical relationships. However, the authors have reduced the problem specifically in hopes that baseline performance will suffer, leaving more room for additional assumptions to improve the amount of learning we are able to do, as well as to increase confidence that the assumption itself will hold.

\subsection{Constraining Assumption of Reflection}
\label{approach:assumption}

Our approach follows the current baseline \cite{livio-baldini-etal-2019} by training a transformation $f_\theta$ with the goal of transforming examples with the same entity-pairs $s_1, s_2$ in a similar way based on $x$, i.e so their dot products are positive when entities match and 0 when they don't, as shown in Equation \ref{eq:dot-product-pos} where $\delta_{ab}$ is the Kronecker delta (1 iff. $a$ matches $b$). However, in order to cope with the reduced semantic space, we propose a new assumption that will potentially allow us to learn more from the dataset: that of inverse relationships. We define an inverse relationship as any relationship statement where the subject and object are reversed, i.e: $\delta_{s`o}\delta_{o`s}$, and assume that they should be represented as the \textit{negative} of the original relationship, as shown in Equation \ref{eq:dot-product-inv}. This can be effectively interpreted as a 'reflection operation' on the relationship, which inverts both the character pair and the relationship vector itself. Thereby, the assumption can be restated as the relationship holding under a reflection operation.

\begin{equation}
    \label{eq:dot-product-pos}
    f_\theta(E)\cdot f_\theta(E`) = 1\ |\  \delta_{s`s}\delta_{o`o} = 1
\end{equation}

\begin{equation}
    \label{eq:dot-product-inv}
    f_\theta(E)\cdot f_\theta(E`) = -1\ |\  \delta_{s`o}\delta_{o`s} = 1
\end{equation}

This may seem like a very common-sense assumption about relationships in general. It's useful to reduce it to a more basic one: relative entity representations should stay fairly constant when being defined as either subject or object. Given this, with two vectors for entity representations $A, B$, and constant context-shift $E$, it's easy to see that $A-B = -((B+E)-(A+E))$, i.e relationships hold under reflection. Our model will take advantage of this to aid its training, and will essentially look to optimize each example's context-shift $\widehat E$ to make it constant over the entire example. Note that doing this successfully in turn implies that the relationship is being successfully separated from the context -- we can conclude that this assumption should aid our ability to perform the task.

\subsection{Task Definition}
\label{approach:task}

Drawing on previous work, we use a modified cross-entropy error calculation to describe whether two relation statements $r$ and $r`$ genuinely describe the same relation when transformed by $f_\theta$, defined more closely as Equation \ref{eq:prob-relations-same}. In addition to this, we offer a related equation to handle the inverse case, shown as Equation \ref{eq:prob-relations-inv}.

\begin{equation}
    \label{eq:prob-relations-same}
    \small{l_{r`r} = p(l=1|r,r`)=\frac1{1+\exp f_\theta (r)^Tf_\theta(r`)}}
\end{equation}

\begin{equation}
    \label{eq:prob-relations-inv}
    \small{l_{-r`r} = p_{inv}(l=1|r,r`)=\frac1{1+\exp( -f_\theta(r)^Tf_\theta(r`))}}
\end{equation}

In order to complete our definition of the new unsupervised approach, we propose training against a new, modified loss function presented as Equation \ref{eq:inv-loss}. This loss function is a modification of a baseline cross-entropy loss, scoring model results by effectively splitting examples into three groups and evaluating performance.

\begin{equation*}
    \label{eq:inv-loss}
    \begin{split}
    L(E) = -\frac1{|E|^2} \sum_{\langle s, o, r \rangle \in E}\sum_{\langle s`, o`, r` \rangle \in E} \\ \delta_{s`s}\delta_{o`o} \cdot \log l_{r`r} + \delta_{s`o}\delta_{o`s}\cdot\log l_{-r`r} + \\ (1-\delta_{s`s}\delta_{o`o})(1-\delta_{s`o}\delta_{o`s}) \cdot \log(1-l_{r`r})
    \end{split}
\end{equation*}

In order to perform this calculation in a feasible amount of time, we follow the literature by performing noise-contrastive estimation to select examples from each of the three groups. In our case, this step has to be adjusted to include a new group of 'hard inverse' examples, if such exists for the current character pair.

\subsection{Evaluation Methodology}
\label{approach:eval}

Although the baseline model was originally evaluated by looking at its predictive power and confusion via looking at N-way K-shot predictions, the authors also discussed the potential for relationship clustering and the comparison to real-world data. This paper, however, \textit{focuses} on this aspect as its downstream task, and so the authors want to check whether the model is picking up any relationship 'archetypes'. In order to evaluate performance on these downstream tasks without explicitly implementing those, we introduce an evaluation method that involves a comparison to human annotations, thereby effectively providing a check for the details learned by the models. This evaluation method isn't new and has been discussed by \citet{schnabel-etal-2015-evaluation}, who would refer to it as Categorization and offers more examples of its use in literature.

We will ask 3 human evaluators who are more familiar with the dataset and its source material to look at the top 36 most-prolific character pairs, i.e those with the most amount of examples attributed to them, and cluster them into some number of unique categories based on prior knowledge and the examples themselves (which were then reconciled by the author, with high-contention examples excluded). We decided to instruct participants to make 6 clusters, to follow a common $\sqrt{n}$ rule-of-thumb. These will be our ground truth clusters. To evaluate the embeddings generated by our transformer model, we will attach these cluster labels to each of the examples and calculate an aggregate Silhouette Coefficient.

The Silhouette Coefficient is defined, very broadly, as in Equation \ref{eq:silh}, where: $a$ is the mean intra-cluster distance and $b$ is the mean nearest-cluster distance. It generally ranges from -1 to 1, with higher scores being indicative of a group that's very well defined, i.e has a high mean distance to other clusters and a low mean distance from points within its own cluster. Therefore, the best embedding method should have higher scores overall. Since these embeddings are built with dot products, magnitudes might confuse scoring so we will use cosine similarity.

\begin{equation}
    \label{eq:silh}
    S = \frac{b - a}{\max(a, b)}
\end{equation}

Note that the transformer is being evaluated on data it has partially seen during training. This is acceptable because our downstream task is not to create a generic relationship extraction transformer but one suited for extraction on a specific text. By evaluating the model in this way, we will be able to (a) compare it to actual semantic labels we have confidence in, and (b) decide whether or not an important aspect of our downstream tasks can be achieved: the discovery of similar relationships.

%% file: tex/experiments.tex
\section{Experiments}
\label{sec:experiments}

In this section, the author fine-tunes two distilled BERT models provided by the HuggingFace library, one as a baseline proposed by previous work, and one with the adjusted loss function taking advantage of the assumptions made, as proposed in section \ref{approach:embed}.

\subsection{Data}

As previously described in Section \ref{approach:data}, our approach utilizes publicly-released scripts for 4 TV Shows and 2 movies from the \textit{Star Trek} franchise, compiled and formatted by \citet{broughton-2018}. This dataset has yet to be embraced by the literature, despite offering several advantages to the field of narrative-relationship extraction, which include:

\begin{enumerate}
    \item The formatting of the dataset allows for labeling and extraction of full character names, with 100\% accuracy. This stands in contrast to the application of NER methods in previous approaches which operate on established datasets compiled from larger corpora such as Wikipedia -- a NER character-extraction process has to be automated and cannot be fully trusted during training on a large, unlabeled corpus.
    \item The closed nature of the source material's plot and the TV-show format make this data well-suited as a proxy for the narrative-relationship extraction task. The franchise is known to keep relationships between characters simple, while also including some evolution between central characters, thereby providing lots of generic archetypes for the model to discover. Similarly, the TV Show format allows for each relationship to appear multiple times over the span of a story life-cycle within each episode, allowing for a more complete representation.
\end{enumerate}

For training, the dataset uses an 80-20 split. The author wants to ensure that there aren't many unseen entity-pair labels in the testing split, so we filter out relations that only have 4 or fewer examples. This results in the distribution seen in Table \ref{table:data-stats}.

\begin{table}[!hbtp]
\centering{
\begin{tabular}{ crr }
    \textbf{Set} & \textbf{Count} & \textbf{Mean Length} \\
    \hline
    \hline
    Train & 63191 & 43 \\
    Test & 15798 & 41 \\
    \hline
    \hline
    Total & 78989 & 42 \\
    Validation & 2683 & 41 \\
\end{tabular}
}
\caption{\label{table:data-stats} Distribution of dataset splits, including the number of examples in each and the mean example tokens length. Note the validation set consists of examples from the top 34 most prolific pairs, and its examples will be a mixture of training and testing data.}
\end{table}

Because of the unfeasibility of computing a full loss function on the dataset even of small size, this dataset is sub-sampled during training using noise-contrastive estimation, according to the method described by \citet{livio-baldini-etal-2019} in their training process. The proposed method also involves sub-sampling to get 'hard-inverse' examples, as described in Section \ref{approach:embed}. Although previous work has not expanded on their methodology, our baseline model keeps the ratio between pos/neg at 1 and strives to keep the ratio between pos+inv/neg at 1 as well. While the sampling method we propose here is absolute, we perform no hard checks for performance reasons due to the possibility of inverse examples simply not existing -- our assumptions let us maintain the proper ratio for 98.2\% of the data examples. Our training uses a hyper-parameter of 8 samples for each example in the batch

\subsection{Models}

Our models utilize the standard BERT tokenizer provided by HuggingFace. This tokenizer had the aforementioned special Subject/Object marker tokens added, and the models had their first layer adjusted for this increase in vocab size.

\begin{figure}[htbp!]
\centering
\includegraphics[width=\columnwidth]{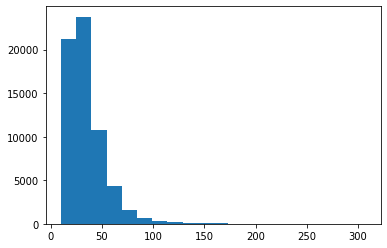}
\caption{Graph of distribution of length of tokens over the full dataset of examples.}
\label{length_distribution}
\end{figure}

The tokenization process also involved truncating the data to limit the size of the input to 104 -- this value allows us to keep 99\% of the original examples untouched, as visible from the distribution of example length in Figure \ref{length_distribution}.

The baseline model is a reproduction of the approach offered by \citet{livio-baldini-etal-2019}, with distil-BERT swapped in for their original model. The training process is optimized to give similar representations to matching entity pairs and dissimilar to non-matching pairs, according to a custom loss function described in their paper. This model takes initial weights from a match-the-blanks pre-training and was referred to by their paper as $\mathrm{BERT}_{EM}\mathrm{+MTB}$, here renamed to $\mathrm{BERT}_{EM}$ for brevity. Because of the aforementioned changes, changes in the dataset, and the new proposed validation method, $\mathrm{BERT}_{EM}$ results have been recomputed to be comparable to the results of this paper.

The proposed alternative model has been similarly initialized from a distil-BERT trained on match-the-blanks tasks, but optimizes an alternative loss function and uses alternative negative/positive selection as described in Section \ref{approach:embed} (i.e training to only ensure relationship for A-B is inverse of B-A). It's referred to as $\mathrm{BERT}_{Inv-}$ in this paper, signifying the assumption of negative inverse relationships. We do not test a hypothetical $\mathrm{BERT}_{Inv+}$ model.

Both models are using the following learning parameters, originally used by the baseline paper and confirmed by a short tuning round:

- Learning Rate: 3e-5 with AdamW

- Batch Size: 32

- Epochs: 5

\begin{figure}[!hbtp]
    \centering
    \includegraphics[width=\columnwidth]{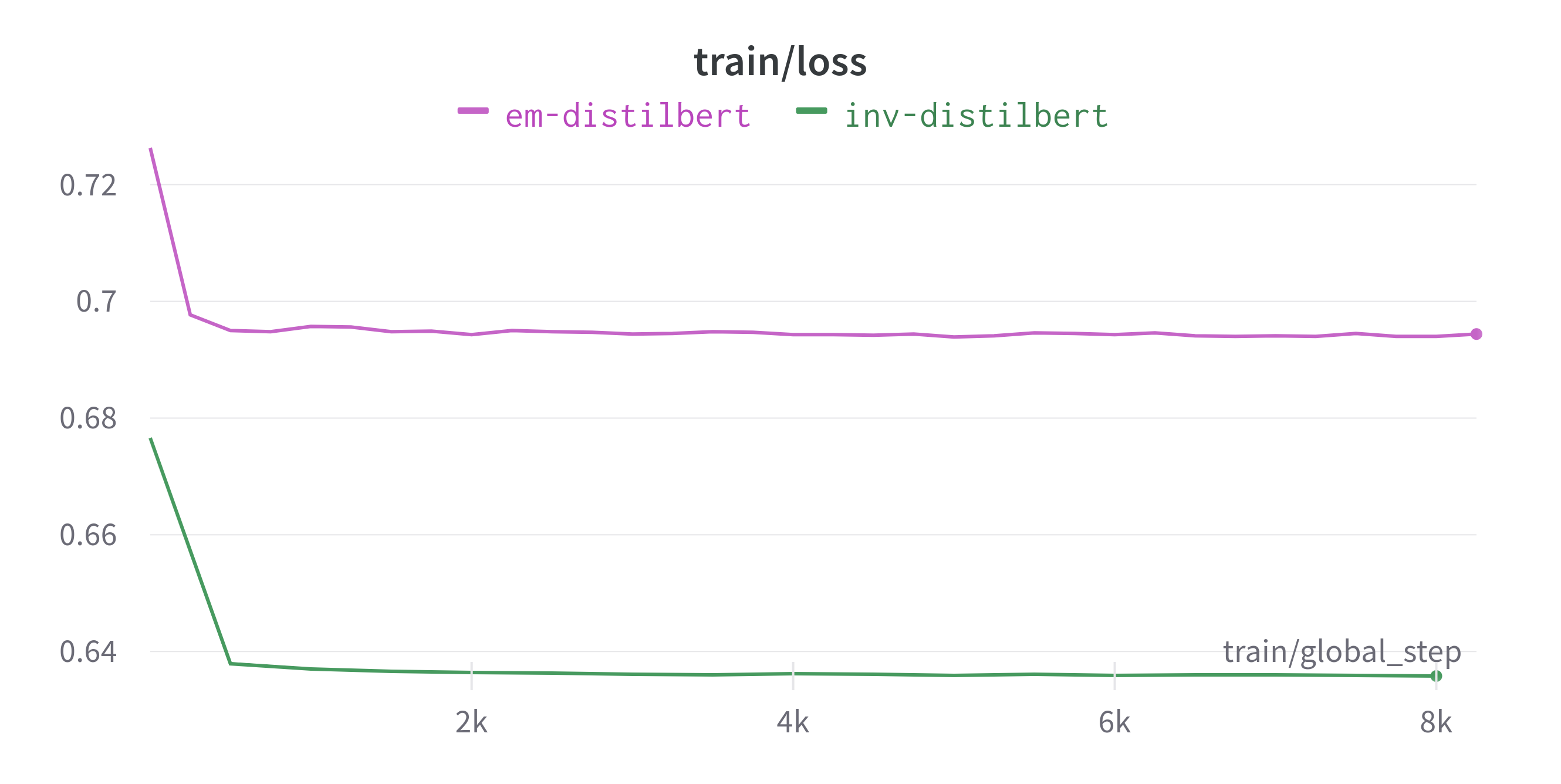}
    \caption{Training Loss for the $\mathrm{BERT}_{EM}$ and $\mathrm{BERT}_{Inv-}$ Model. Note these are not directly comparable.}
    \label{fig:em_loss}
\end{figure}

Training took ~7 hours for the baseline model and ~9 hours for the proposed new approach. The train/test loss for the baseline is shown in Figure \ref{fig:em_loss}, with the best epoch for both models highlighted in Table \ref{tab:loss}.

\begin{table}[!hbtp]
    \centering
    \begin{tabular}{c||c|c|c}
        Model & Train & Test & Epoch \\
        \hline
        {\footnotesize $\mathrm{BERT}_{EM}$} & 0.6943 & 0.7616 & 2 \\
        {\footnotesize $\mathrm{BERT}_{Inv-}$} & 0.6362 & 0.6465 & 2 \\
    \end{tabular}
    \caption{Training and Testing Loss for 'best' epoch with lowest Testing Loss score. Note these are not directly comparable due to modified loss.}
    \label{tab:loss}
\end{table}

\subsection{Evaluating Models on Ground-Truth Clusters}

As previously described in Section \ref{approach:eval}, our evaluation method involves computing silhouette scores for ground-truth clusters from human participants. These scores were computed by using the Validation dataset as described in Table \ref{table:data-stats}, which is comprised of all the examples belonging to a labeled subject/object pair, including examples used both in training and testing previously. Two examples were removed due to contention (<ARJIN, DAX> and inverse), leaving 34 examples. This data is passed through the model to generate relationship embeddings with an associated group label for each example.

The relationship embeddings were scored via the cosine-similarity silhouette method in three ways: (a) by scoring across all examples with labels that put each character-pair in their own group; (b) by scoring across all examples with labels that put each character-pair in a ground-truth clustering; and (c) by scoring across the composite (averaged across character-pair examples) relations, with labels that put each character-pair into their ground-truth clustering. The results of this are shown in Table \ref{tab:silhouette}.

\begin{table}[!hbtp]
    \centering
    \begin{tabular}{c||c|c}
        Model ($\sigma$) & $\mathrm{BERT}_{EM}$ & $\mathrm{BERT}_{Inv-}$ \\
        \hline
        Character & 0.172 \footnotesize{(1.99e-2)} & -0.045 \footnotesize{(1.32e-2)} \\
        Cluster & -0.049 \footnotesize{(0.98e-2)} & -0.247 \footnotesize{(8.42e-2)} \\
        Composite & -0.227 \footnotesize{(5.38e-2)} & -0.084 \footnotesize{(4.65e-2)} \\
        
    \end{tabular}
    \caption{Silhouette Coefficient for examples by Character-Pair, Ground-Truth, and Composite Relationships,  given embeddings generated by models. Generally, the coefficient ranges from -1 to 1, with negative values usually indicating some overlap between clusters; higher is better.}
    \label{tab:silhouette}
\end{table}

The scores shown in Table \ref{tab:silhouette} clearly indicate that the baseline SOTA model $\mathrm{BERT}_{EM}$ outperformed the proposed $\mathrm{BERT}_{Inv-}$ by a wide margin on all but the composite clustering task. Notably, the character clusters are much more defined, reflecting the training target. Despite this, both models show some degree of separation of the clusters, meaning that the relationship 'archetypes' \textit{are actually} being picked up by both models. Furthermore, as we will show in Section \ref{sec:analysis}, the proposed model is actually superior on some of the categories presented here, as hinted by the composite score in Table \ref{tab:silhouette}.

%% file: tex/analysis.tex
\section{Analysis}
\label{sec:analysis}

\begin{table*}[!htbp]
\centering\small 
\begin{tabular}{c||c|c|c|c|c|c}
    \bf{Cluster \#} & \multicolumn{1}{c}{\bf{1}} & \multicolumn{1}{c}{\bf{2}} & \multicolumn{1}{c}{\bf{3}} & \multicolumn{1}{c}{\bf{4}} & \multicolumn{1}{c}{\bf{5}} & \multicolumn{1}{c}{\bf{6}} \\
    \hline
    Proportional Size & 0.353 & 0.176 & 0.118 & 0.176 & 0.059 & 0.118 \\
    \hline
    $\mathrm{BERT}_{EM}$ Clust. & -.1042 \tiny{(2.33e-3)} & .0457 \tiny{(7.49e-3)} & -.0124 \tiny{(4.89e-3)} & -.0953 \tiny{(3.51e-3)} & -.1411 \tiny{(3.04e-3)} & .0801 \tiny{(8.56e-3)} \\
    $\mathrm{BERT}_{Inv-}$ Clust. & -.4350 \tiny{(2.41e-3)} & .1494 \tiny{(14.0e-3)} & .0140 \tiny{(4.33e-3)} & -.4978 \tiny{(2.69e-3)} & -.5742 \tiny{(6.11e-3)} & .1044 \tiny{(12.1e-3)} \\
    \hline
    $\mathrm{BERT}_{EM}$ Comp. & -.3145 \tiny{(7.28e-3)} & -.0586 \tiny{(28.2e-3)} & -.2412 \tiny{(37.3e-3)} & -.3447 \tiny{(19.1e-3)} & -.6193 \tiny{(.046e-3)} & .1714 \tiny{(18.8e-3)} \\
    $\mathrm{BERT}_{Inv-}$ Comp. & -.1125 \tiny{(11.1e-3)} & .0488 \tiny{(46.2e-3)} & -.3062 \tiny{(36.9e-3)} & -.1779 \tiny{(2.59e-3)} & -.1062 \tiny{(84.7e-3)} & .1775 \tiny{(111e-3)} \\
    \hline
    Improved by Comp. & \tiny{$\mathrm{BERT}_{Inv-}$} & None & None &  \tiny{$\mathrm{BERT}_{Inv-}$} & \tiny{$\mathrm{BERT}_{Inv-}$} & Both
\end{tabular} 
\caption{Silhouette scores and variance for each ground-truth cluster with size given for reference. The first group of results measures scores of individual examples in each cluster, while the second group measures scores of Composed examples (embeddings averaged by character). The impact of composition is highlighted in the last row.}
\label{tab:cluster_errors}
\end{table*}

Although the proposed model fails to make the ground-truth clustering significantly more sensible, it demonstrates improvement in the clusters where our assumption is more likely to hold, as shown when the errors are split in Table \ref{tab:cluster_errors}, as well as demonstrating broad improvement on composed relationship embeddings. Furthermore, some broad, unique trends are visible in the table, which gets analyzed more closely in the following sections.

\subsection{Impact of the Composition Step}
\label{analysis:composition}

Notably from the baseline $\mathrm{BERT}_{EM}$ results, we can see that composed relationships, made by averaging all the relationship embeddings for a character, score worse than when examples are separate, with the exception of Clustering 6 seen in Table \ref{tab:cluster_errors}. In general terms, the authors attribute this to differences of bias / variance trade-offs between the model predictions.

The mechanism for this has simple causes: the aggregation step is masking a lot of 'outlier' points. If these points are ones defining the critical region of a character relationship 'archetype' represented by the cluster -- i.e the outliers incidentally overlap with those of related character-pairs -- masking them would remove that dense region of overlap and replace it with more-distant composite points. This in turn \textit{expands} the decision region for the cluster, hurting the score. Alternatively, if these are genuine outliers with less overlap, masking them would define the cluster more tightly, \textit{tightening} the decision region and raising the score. Clusters with high variance would benefit from this step if their bias is low and overlapping regions are not just defined by outliers.

\begin{figure}[htbp!]
    \centering
    \includegraphics[width=\columnwidth]{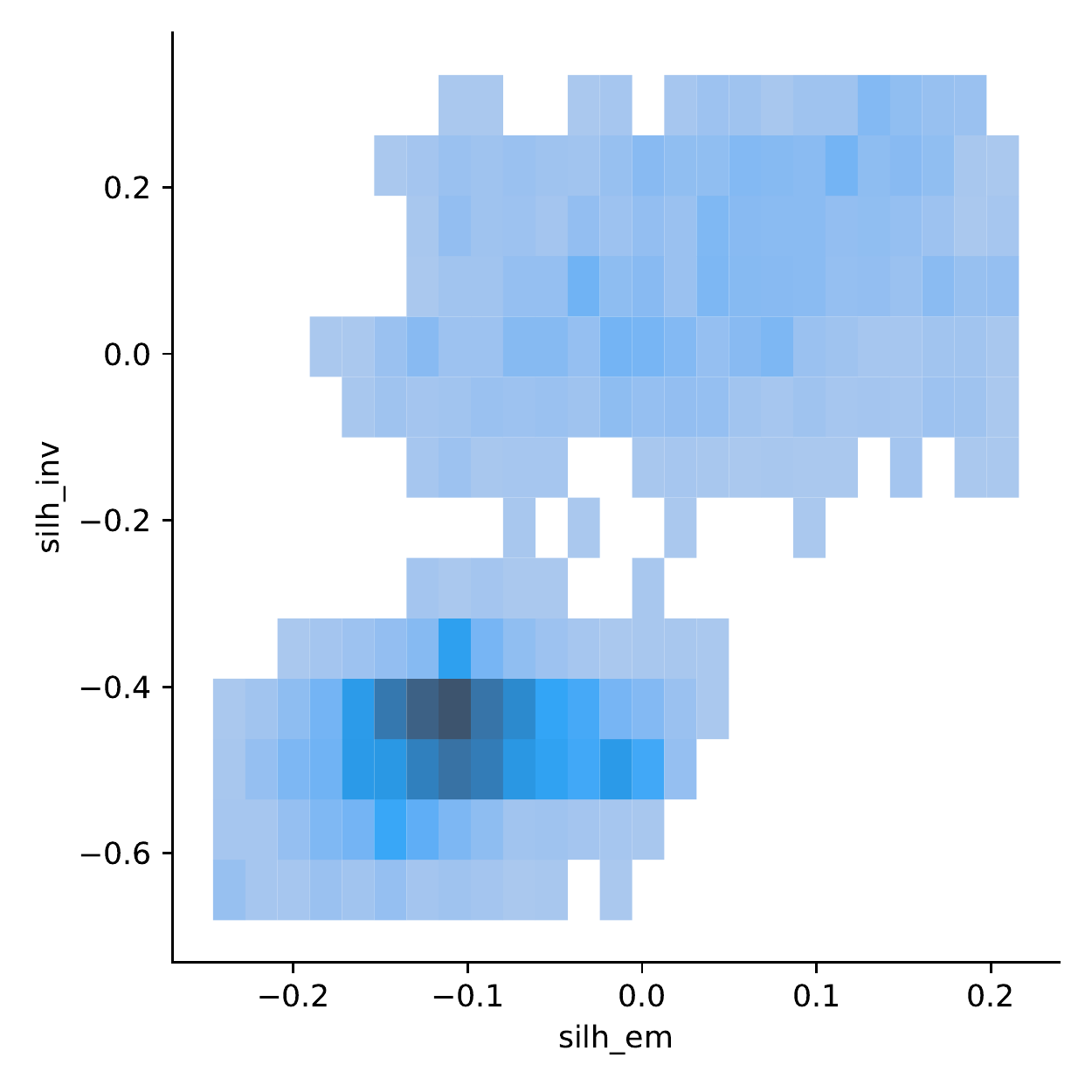}
    \caption{Density of $\mathrm{BERT}_{EM}$ and $\mathrm{BERT}_{Inv-}$ score distributions for individual examples on ground-truth clusters.}
    \label{fig:example-cluster-score-both}
\end{figure}

The dataset is also expected to have a lot of examples that fail to define a relationship, such as ones where characters talk about the weather or generic topics. Every cluster and relationship pair will have high variance with lots of outliers; the way the model defines their distribution will determine how that cluster performs after composition. This 'confusion region' is responsible for some of the dark blue regions at the bottom left in Figure \ref{fig:example-cluster-score-both}.

The 6th cluster is unusual in this case, benefiting from the composition step across both models. The cluster actually contains a dialogue between 4 different alien or android characters, a detail that leads to significant stylistic differences in some of the dialogue examples and that makes them easier to group. From the elevated composition scores, we can tell that its definition is not just incidental but is in fact 'stable', i.e supported by a majority of their points instead of a minority or outliers, and generally maintains high accuracy.


\begin{figure}[htbp!]
    \centering
    \includegraphics[width=\columnwidth]{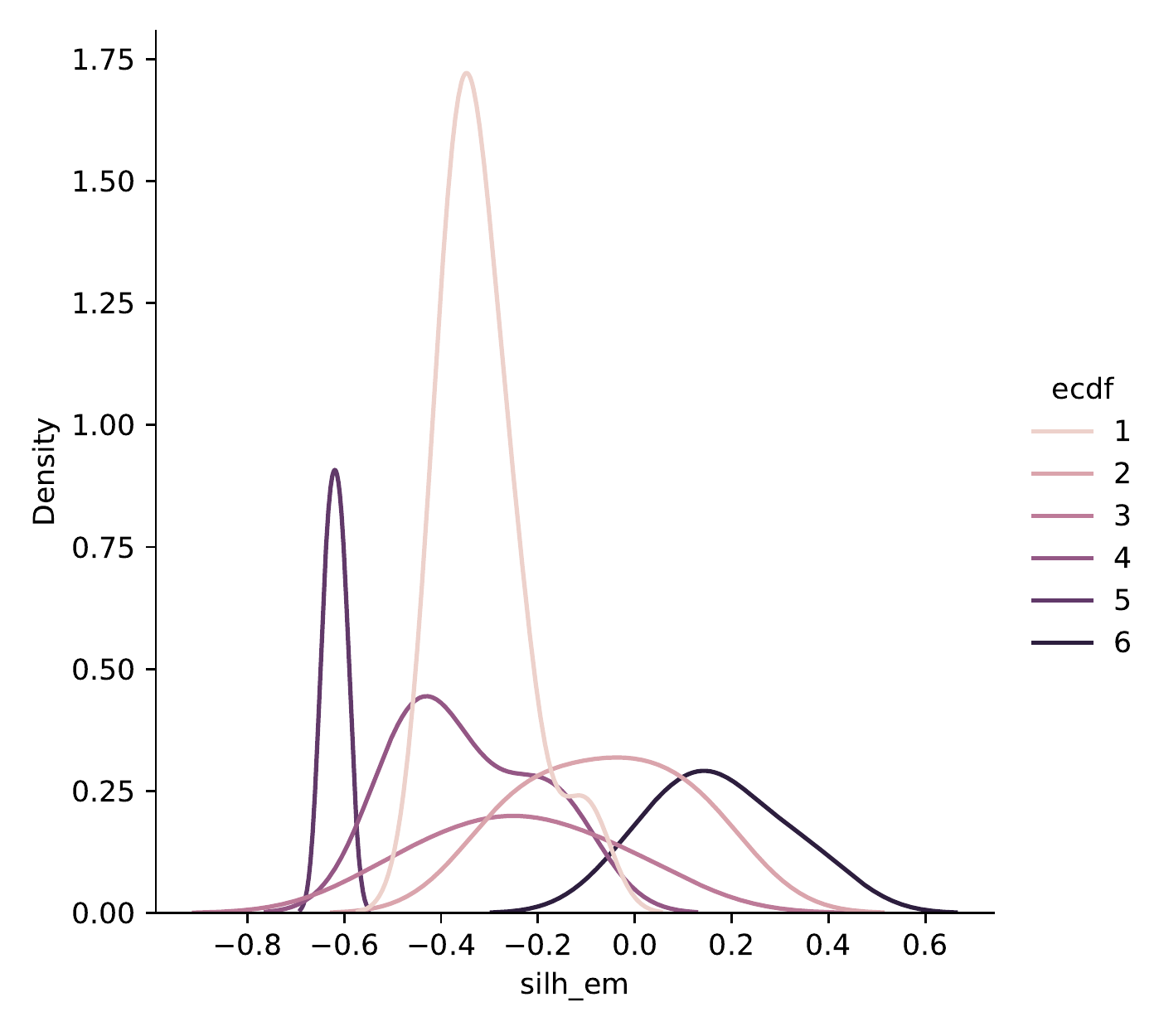}
    \includegraphics[width=\columnwidth]{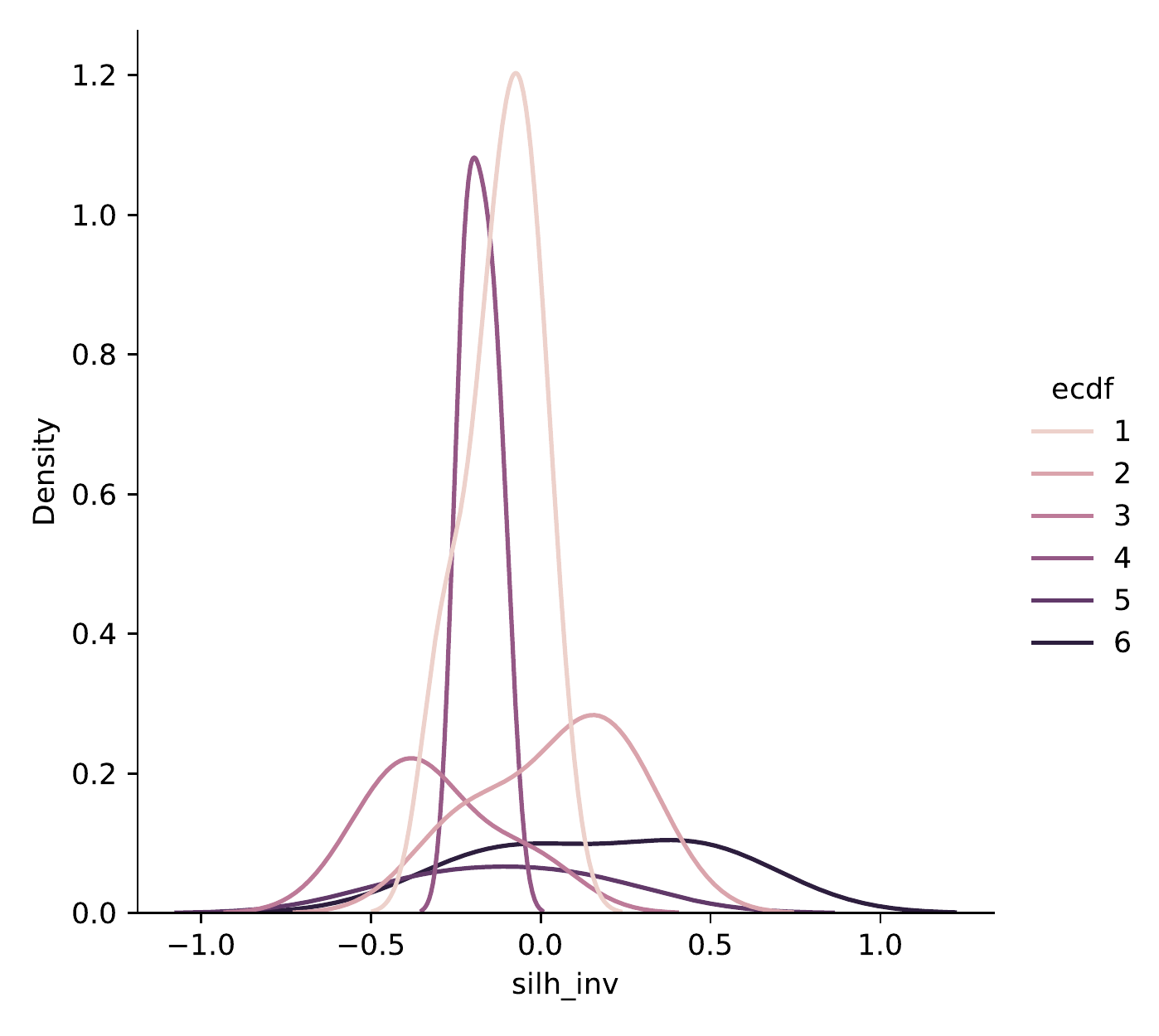}
    \caption{$\mathrm{BERT}_{EM}$ and $\mathrm{BERT}_{Inv-}$ score distributions for aggregate pair relationships on ground-truth clusters, split by cluster. Rightmost is better.}
    \label{fig:comp-cluster-score}
\end{figure}

\subsection{Analyzing Differences in Success and Failure}
\label{analysis:differences}

\begin{figure}[htbp!]
    \centering
    \includegraphics[width=\columnwidth]{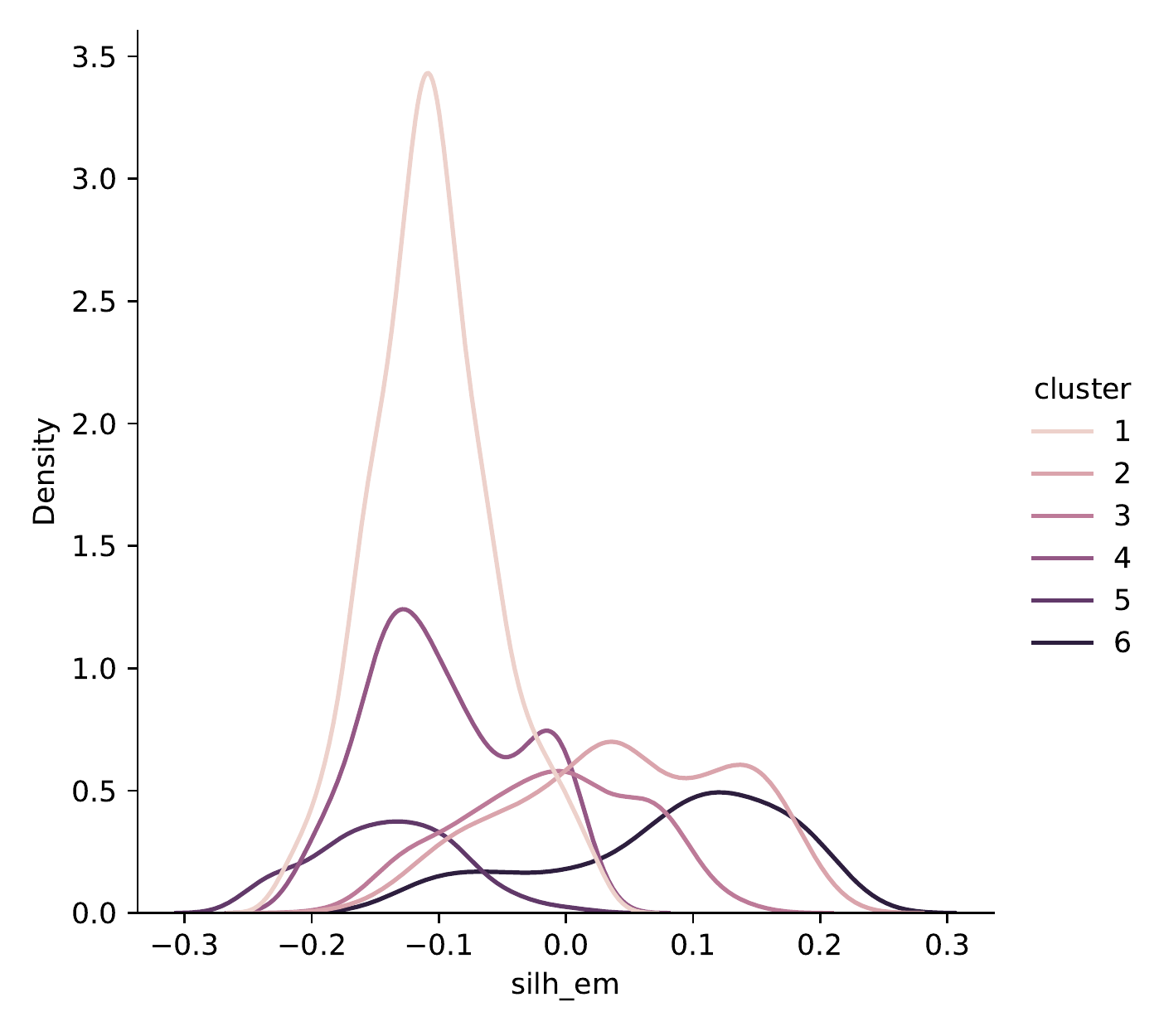}
    \includegraphics[width=\columnwidth]{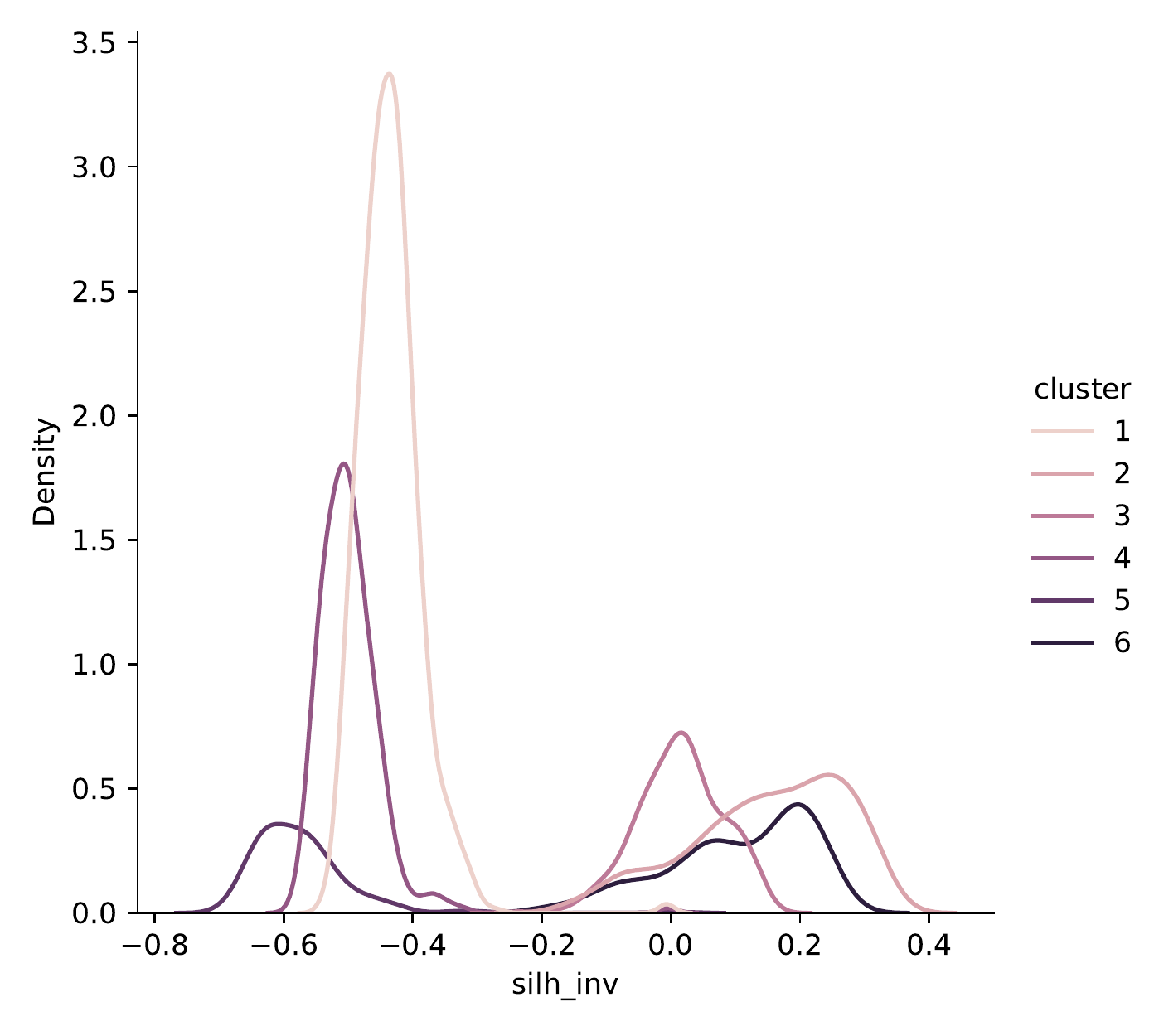}
    \caption{$\mathrm{BERT}_{EM}$ and $\mathrm{BERT}_{Inv-}$ score distributions for individual examples on ground-truth clusters, split by cluster. Rightmost is better.}
    \label{fig:example-cluster-score}
\end{figure}

It is useful to look further into the differences on an example level, without aggregation. Both models seem to perform similarly when evaluated this way: as evidenced by Figure \ref{fig:example-cluster-score}, they provide similar distributions for the same clusters: 2, 3, and 6, while confusing clusters 1, 4, and 5. This result is actually hinted at by the human labels: multiple examples were omitted from the same failing clusters due to contention, potentially indicating that they are semantically similar. As evidenced by example scores in Figure \ref{fig:example-cluster-score}, the baseline $\mathrm{BERT}_{EM}$ is able to handle this similarity better at an example level.

The stark bimodal distribution of scores for the proposed $\mathrm{BERT}_{Inv-}$ model is a result of this issue specifically: examples from these failing clusters have mutually-high confusion rates, potentially due to the aforementioned semantic similarity, resulting in scores around -.5 as seen in Table \ref{tab:cluster_errors}. However, when aggregated by character as in Figure \ref{fig:comp-cluster-score}, these scores become superior to baseline results, implying that this issue is not of bias but precision: errors are normally distributed and the model is failing to learn enough details. Examples from this group include relations like <Shannon, Henry> and <Trelane, Kirk>, where the baseline $\mathrm{BERT}_{EM}$ drastically outperforms the proposed model because of how interchangeable the characters are in a majority of their dialogue.

Conversely, the proposed model beats the baseline at classifying the lopsided romantic relationships included in Cluster 2. Among the most successful examples are dialogues from <Odo, Arrissa> (scores of .18 vs .02 for baseline) and include dialogue describing betrayal and romance, allowing for both personalities to be apparent.

Another interesting result is the relationship between Emh and Emh2 from Cluster 6 -- in-universe, these seem to be two navigation androids who provide support for other characters. Their dialogues are generally interchangeable, and while the baseline $\mathrm{BERT}_{EM}$ performs well on some of these, the authors expected this to be consistently confusing for $\mathrm{BERT}_{Inv-}$. The truth is more nuanced -- the proposed model is able to group pieces of dialogue with a clear power difference between the robots (scores of .2 for dialogue where Emh2 is serving Emh, while the baseline scores -.1), but fails where the dialogue becomes more equal and the characters more interchangeable. The models seem to have inversely-correlated scores on this character pair, again suggesting that the proposed model might favor dialogue with distinguishable characters, while the baseline might be paying more attention to the unique style of the majority of examples.


\subsection{Limitations of Methodology and Model}
\label{analysis:limits}


Previous iterations of this work evaluated the embeddings with silhouette scores based on Euclidean L2 distance instead of Cosine distance -- this is the incorrect way to look at the embeddings due to the models optimizing against embedding dot products. This change should be noted however because the change also moved the most-defined cluster (5th) down to the least-defined, lowest-scoring one. The authors cannot point to a reason for this decrease aside from the high variance of results inherent in a tiny cluster such as this (2 examples). This points to one particular weakness of using Silhouette Scores to evaluate embeddings: failure to define one cluster will significantly impact the scores of others. Fortunately, the results reported in Table \ref{tab:cluster_errors} are fairly stable when the clusters are randomly down-sampled, so interference is minimal in the reported results.

Furthermore, due to time and cost constraints, these models were both trained on a reduced-size, distilled model, which would significantly impact their learning ability. The unusual dataset could also fail to represent the task of narrative relationship extraction accurately, leading to a loss of generalizability. Most notably, however, due to the author's failure to find code associated with the baseline approach provided by \citet{livio-baldini-etal-2019}, a significant portion of the time had to be spent replicating their model approach. Our reproduction likely failed at accurately representing all aspects of it, especially including the unique sampling methodology, and the efficiency of training, all of which would affect the rate of learning and impact our results. Our implementation is released in Appendix \ref{sec:appendix}, and can hopefully provide a baseline for future work.

%% file: tex/conclusion.tex
\section{Conclusion and Future Work}
\label{sec:conclusion}

In this paper, we have presented a novel modification to the SOTA approach for relationship extraction by including an assumption specific to the narrative category, which is: relationships between characters are inverted when characters are switched, i.e they hold under the reflection operation. Although the model we built failed to outperform the baseline in absolute terms, we showed that it works well in conditions where that assumption holds and fails where its effects might be less pronounced, i.e for relationships where characters might be more interchangeable. In addition to this, we've applied a unique method of evaluation on relationship embeddings generated by both baseline and proposed models, which acts as a proxy for downstream clustering and Knowledge Graph Generation tasks and provides a rational-basis check on both models due to \textit{human-created} ground truth labels.

Future work could expand on this research by trying to replicate results with a BERT Large model, as is more common in related literature, as well as removing other limitations discussed in Section \ref{analysis:limits}. More insight is needed into the difference in errors generated by the models discussed here, as it has been shown they are representing these differently and the success of an example differs drastically between them. In addition, the size of ground-truth labeled data could be increased in terms of both size, confidence, and accuracy to give better insight into model performance. Future work could focus on applying models trained with this assumption to evaluate performance on actual downstream tasks, in particular, in Knowledge Graph construction and relationship clustering/classification. Furthermore, the assumption could be further evaluated by analyzing its inverse (a $\mathrm{BERT}_{Inv+}$ model), or by admitting a wider range of named entities into the process to evaluate it more broadly.

%% file: tex/acknowledgements.tex
\section*{Acknowledgements}

We'd like to acknowledge Dr. Jinho Choi of Emory University for his supplemental advice, as well as our 3 human participants involved with labeling the data: Sarah Harari Minian, and two users who asked to remain anonymous.

%% file: tex/appendix.tex
\section{Appendix}
\label{sec:appendix}

We have released the implementation details of both of these models, as well as their trained forms publicly. 

\begin{enumerate}
    \item The pre-trained $\mathrm{BERT}_{Inv-}$ model can be found at \url{https://huggingface.co/mfigurski80/relation-distilbert-inv}.
    \item The pre-trained $\mathrm{BERT}_{EM}$ model can be found at \url{https://huggingface.co/mfigurski80/relation-distilbert-em}.
    \item The tokenizer used for both of these models can be found at \url{https://huggingface.co/mfigurski80/narrative-relationship-tokenizer}.
    \item The code behind constructing and evaluating these models is publicly released at \url{https://github.com/mfigurski80/rel-extraction}
\end{enumerate}